%% file: Template.tex
\documentclass{article}
\usepackage{spconf,amsmath,graphicx,hyperref}
\usepackage{amssymb}  
\usepackage{booktabs}
\usepackage{multirow}
\usepackage{caption}
\usepackage[normalem]{ulem}
\usepackage{subcaption}
\usepackage{etoolbox}
\usepackage{makecell}

\makeatletter
\patchcmd{\thebibliography}{\settowidth\labelwidth{[#1]}}{\settowidth\labelwidth{[#1]}\setlength{\itemsep}{0pt}\setlength{\parsep}{3.2pt}}{}{}
\makeatother

\title{TIPS Over Tricks: Simple Prompts for Effective Zero-Shot Anomaly Detection}
\usepackage{xcolor}
\usepackage[table]{xcolor}

\name{
\itshape
\centerline{%
\begin{minipage}{0.9\linewidth}\centering
Alireza Salehi$^{1}$, Ehsan Karami$^{1,*}$, Sepehr Noey$^{2,*}$, Sahand Noey$^{2,*}$, \\
Makoto Yamada$^{3}$, Reshad Hosseini$^{1}$, Mohammad Sabokrou$^{3}$%
\thanks{* Equal Contribution}
\end{minipage}%
}
}

\address{
\small
$^{1}$University of Tehran\hspace{0.8cm}
$^{2}$Amirkabir University of Technology\hspace{0.8cm}
$^{3}$Okinawa Institute of Science and Technology
}

\newcounter{paranum}[section]

\newcommand{\methodname}{\texttt{Tipsomaly} }
\newcommand{\methodnamec}{\texttt{Tipsomaly}, }

\newcommand{\fixedprompt}{P_f}  
\newcommand{\fixedfeatures}{G_f}  

\newcommand{\globalobject}{g_o^i}  
\newcommand{\globalspatial}{g_s^i}

\newcommand{\learnprompt}{P_l}  
\newcommand{\learnfeatures}{G_l}

\begin{document}

%
\maketitle


\input{sec/abstract}

\vspace{-1.5em}
\input{sec/intro}


\input{sec/method}

\input{sec/experiments}

\vfill\pagebreak

\section{Acknowledgements}
\vspace{-1em}
Mohammad Sabokrou’s work in this project was supported by JSPS KAKENHI Grant Number 24K20806.

\vspace{-1em}
\begingroup
\small
\setlength{\baselineskip}{9pt}
\bibliographystyle{IEEEbib}
\bibliography{strings}
\endgroup

\section{Appendix}
\input{sec/appendix}

\end{document}

%% file: sec/abstract.tex
\begin{abstract} 
Anomaly detection identifies departures from expected behavior in safety-critical settings. When target-domain normal data are unavailable, zero-shot anomaly detection (ZSAD) leverages vision–language models (VLMs). However, CLIP’s coarse image–text alignment limits both localization and detection due to (i) spatial misalignment and (ii) weak sensitivity to fine-grained anomalies; prior works compensate with complex auxiliary modules yet largely overlook the choice of backbone. We revisit the backbone and use TIPS—a VLM trained with spatially aware objectives. While TIPS alleviates CLIP’s issues, it exposes a distributional gap between global and local features. We address this with decoupled prompts—fixed for image-level detection and learnable for pixel-level localization—and by injecting local evidence into the global score. Without CLIP-specific tricks, our TIPS-based pipeline improves image-level performance by 1.1–3.9\% and pixel-level by 1.5–6.9\% across seven industrial datasets, delivering strong generalization with a lean architecture. Code is available at \href{https://github.com/AlirezaSalehy/Tipsomaly}{\texttt{github.com/AlirezaSalehy/Tipsomaly}}.
\end{abstract}

\begin{keywords}
Zero-shot Anomaly Detection, Vision-language Model
\end{keywords}

%% file: sec/intro.tex
\section{Introduction}
\vspace{-0.5em}
\label{sec:intro}
Anomaly detection seeks rare deviations in visual data and is pivotal in safety-critical settings such as industrial inspection and medical imaging~\cite{unifiedsurvey,esmaeili2023generative}. Conventional approaches commonly assume access to sufficient in-domain normal data and often model it via deviations in feature distribution~\cite{mkd,sabokrou2018adversarially, sabokrou2018deep,sabokrou2020deep} or reconstruction~\cite{sabokrou2016video,hu2024anomalydiffusion}. However, when labeled in-domain normal samples are unavailable—due to privacy or scarcity—zero-shot anomaly detection (ZSAD) becomes essential. Recent vision–language models (VLMs) such as CLIP~\cite{radford2021learning} make this setting practical by coupling text prompts with robust transferable visual representations, enabling anomaly detection in unseen domains without target-domain training data~\cite{jeong2023winclip, gu2024anomalygpt}.

However, CLIP’s pretraining poses a bottleneck for ZSAD. Its contrastive objective does not enforce patch-level alignment—neither among patch embeddings nor between patches and text~\cite{radford2021learning}. This weak spatial coherence and patch–text grounding lead to degraded localization of fine-grained cues~\cite{clipsurgery}, which in turn hampers both detection and pixel-level performance~\cite{deng2023anovl}. 

To mitigate these issues, prior ZSAD methods introduce non-learnable refinements (``VV'' attention~\cite{zhou2023anomalyclip, deng2023anovl}) and learnable modifications to the vision encoder (trainable visual prompts~\cite{cao2024adaclip}, feature adapters~\cite{chen2023april}). Although helpful, these adaptations cannot fully restore the spatial alignment and patch–text correspondence absent in CLIP inherently, and increase architectural complexity. Moreover, previous analyses suggest that learnable adaptations can overfit train data, weakening the cross-domain generalization required for zero-shot anomaly detection~\cite{salehi2025crane}. For example, AdaCLIP’s~\cite{cao2024adaclip} joint tuning of text and vision encoders can improve image-level detection, but transfers poorly to pixel-level localization, underscoring the limits of CLIP-centric adaptations. 

In this paper, we revisit an underexplored lever for ZSAD—the backbone—and investigate TIPS's~\cite{maninis2024tips} capabilities, a more spatially aware model for the task, showing it sidesteps several CLIP limitations and enables a simpler pipeline without complex adaptations. However, TIPS is trained with a general vision–language objective, making direct utilization suboptimal for anomaly detection. We evaluated learnable prompts and fixed textual templates and found a consistent \emph{distributional gap} between local and global features: when learnable prompts are trained with an additional \emph{global} objective (normal vs.\ anomalous classification) alongside the \emph{local} loss, image-level performance improves but pixel-level alignment degrades, indicating a mismatch between local and global anomaly representations.

To address this, we adopt \emph{decoupled prompting}: fixed textual prompts for image-level scoring, and learnable prompts optimized only with a local loss for pixel-level segmentation. We also leverage TIPS’s two global embeddings—the spatial token and the object-centric token—and assess their ability to discriminate anomalies and aggregate local cues; we then complement the global score with local evidence. Our framework \methodname improves state-of-the-art image-level performance by 1.1–3.9\% and pixel-level metrics by 1.5–6.9\% across 14 industrial and medical datasets.

\begin{figure*}[t]    
    \centering
    \includegraphics[width=1\linewidth]{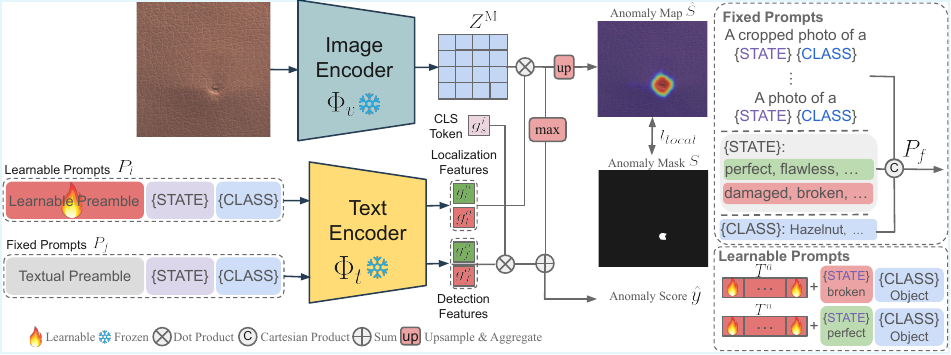} 
    \vspace{-2.2em}
    \caption{ 
    \textbf{Framework of \methodname for zero-shot anomaly detection.}
    An input image is encoded by the TIPS vision encoder $\Phi_v$ into dense patch features $Z^{\mathrm{M}}$ and two global tokens, $\globalobject$ (object-centric) and $\globalspatial$ (spatial).
    We use \emph{decoupled prompting}: the fixed detection prompt set $\fixedprompt$ yields fixed text prototypes $\fixedfeatures$ for image-level decisions, while the learnable localization prompt set $\learnprompt$ produces $\learnfeatures$ for pixel-level localization.
    The anomaly map $\hat{S}$ is computed via patch–text similarities between $Z^{\mathrm{M}}$ and $\learnfeatures$, then upsampled and smoothed to image resolution.
    For image-level detection, spatial token $\globalspatial$ is scored against $\fixedfeatures$, then the result is added to the strongest local evidence, \(\max(\hat{S}_a)\), to obtain anomaly score $\hat{y}$.
    }   
    \label{fig:framework}
\end{figure*}

%% file: sec/method.tex
\section{Method}
\label{sec:method}

\subsection{Problem Definition}
In anomaly detection, a dataset is composed of triplets
$\mathcal{D}=\{(x,y,S)\}$, where each image $x\in\mathbb{R}^{C\times H\times W}$
is paired with an image-level label $y\in\{0,1\}$ and a pixel-level mask
$S\in\{0,1\}^{H\times W}$, with $1$ indicating abnormality.
Let $M_{\theta}$ denote a pretrained vision–language model (e.g., CLIP, TIPS) with parameters $\theta$. In the zero-shot setting, the goal is to detect and localize anomalies in an unseen domain $\mathcal{D}_{\mathrm{unseen}}$ without using any target-domain samples during optimization. To this end, we may adapt the model using a source dataset \(\mathcal{D}_{\mathrm{src}}\) drawn from an auxiliary domain disjoint from \(\mathcal{D}_{\mathrm{unseen}}\) by learning prompts \(P\). At test time, for each $x\in\mathcal{D}_{\mathrm{unseen}}$ the adapted model produces an image-level
anomaly score $\hat{y}(x)\in[0,1]$ and a pixel-level anomaly map \vspace{-0.4em} $\hat{S}(x)\in[0,1]^{H\times W}$:
\[
(\hat{y}(x),\,\hat{S}(x)) \;=\; M_{\theta}(x, P).
\]
Thereafter, the final image- and pixel-level decisions are obtained by thresholding these scores.

\subsection{Feature Extraction}
To produce pixel- and image-level anomaly scores for an image \(x\in\mathbb{R}^{C\times H\times W}\), we extract visual and textual features. As shown in Fig.~\ref{fig:framework}, TIPS—similar to CLIP—comprises a vision encoder \(\Phi_v\) and a text encoder \(\Phi_t\). The vision encoder outputs two global tokens—object-centric \(\globalobject\in\mathbb{R}^{D}\) and spatial \(\globalspatial\in\mathbb{R}^{D}\)—used for image-level decisions, along with dense patch embeddings \(Z^{\mathrm{M}}\in\mathbb{R}^{N\times D}\) for pixel-level localization. The text encoder produces embeddings for normal and abnormal states from two prompt sets: \emph{fixed detection prompts} $\fixedprompt$ aiming for image-level detection and \emph{learnable localization prompts} $\learnprompt$ specializing the text representations for fine-grained localization. This \emph{decoupled prompting} addresses a distributional gap between the global tokens and the patch features of the vision encoder observed in our study.

\subsection{Fixed Detection Prompts}
\label{sec:fix_prompts}
We compose each fixed prompt from three parts: a \emph{template preamble}, a \emph{state phrase}, and the \emph{class token}.
The preamble is drawn from a compact set of generic inspection templates (e.g., \texttt{a cropped photo of a \{\}}),
and the state phrase specifies \emph{normal} vs.\ \emph{abnormal} semantics (e.g., \texttt{flawless \{\}} vs.\ \texttt{damaged \{\}}).
Given a class name \(c\), we instantiate prompts by the Cartesian product
\[
\fixedprompt \;=\; \{\, t(s(c)) \;\mid\; t\in\mathcal{T},\ s\in\mathcal{S} \,\},
\]
and split them into normal and abnormal subsets \(\fixedprompt^{\mathrm{n}}\) and \(\fixedprompt^{\mathrm{a}}\).
Passing each prompt \(p\in\fixedprompt\) through the text encoder \(T(\cdot)\) yields an embedding \(T(p)\in\mathbb{R}^{D}\). We then average embeddings within each subset to obtain fixed text prototypes and denote \(\fixedfeatures = \{g_f^{\mathrm{n}},\,g_f^{\mathrm{a}}\}\). These fixed prototypes are used for \emph{image-level detection}.

\subsection{Learnable Localization Prompts}

We observed that fixed prototypes offer strong image-level alignment but fail to encode local cues needed for patch-level alignment.
Hence, for \emph{localization} we learn two class-agnostic prompt token sets on \(\mathcal{D}_{\mathrm{src}}\),
\(T^{\mathrm{n}}=\big[t^{\mathrm{n}}_{1},\dots,t^{\mathrm{n}}_{E}\big]\) and
\(T^{\mathrm{a}}=\big[t^{\mathrm{a}}_{1},\dots,t^{\mathrm{a}}_{E}\big]\), with \(t^{(\cdot)}_{(\cdot)}\in\mathbb{R}^{D}\).
We denote the resulting localization prompt set by
\[
\learnprompt \;=\; \big\{\, [T^{\mathrm{n}},~\text{``object''}],\; [T^{\mathrm{a}},~\text{``damaged''},~\text{``object''}] \,\big\}.
\]
Passing each element of \(\learnprompt\) through the text encoder \(T(\cdot)\) yields the normal/abnormal
localization prototypes \(\learnfeatures=\{g^n_l,\,g^a_l\}\) which are used to
score patch embeddings \(Z^{\mathrm{M}}\) via similarity.

\subsection{Training Objective \& Inference}
Having obtained the textual embeddings specialized for detection $\fixedfeatures$ and localization $\learnfeatures$, as well as the visual embeddings \( e \in \{ \globalobject, \globalspatial, Z_{j,k} \} \), where \( Z_{j,k} \) denotes the \((j,k)\)-th patch embedding in the unflattened local feature map \(Z^M\), we compute the probability that each visual embedding belongs to the anomalous class, denoted by \(p_a\), by applying the Softmax function to the corresponding similarity scores:

\begin{equation}
p\!\left(e,G_{(\cdot)}\right)=\operatorname{softmax}\!\left(\frac{G_{(\cdot)}^{\top}e}{\tau}\right),
\quad
p_a=\bigl[p(e,G_{(\cdot)})\bigr]_a 
\label{eq:likelihood} 
\end{equation}

where the temperature \( \tau \) is approximately 0.0042 according to TIPS details~\cite{maninis2024tips}. We denote the probability of a visual embedding $e$ being abnormal, $p_a(e, G)$, as its \emph{anomaly score}.

\vspace{0.2em}
\textbf{Training.} We learn the normal/abnormal patch-level visual distribution via learnable prompts $\learnprompt$, while keeping the vision encoder $\Phi_v$ and text encoder $\Phi_t$ frozen and employing local loss functions: Focal loss~\cite{focalloss} and Dice loss~\cite{diceloss}. For output feature map $Z^M$, we compute normal and anomaly maps $\hat{S}_n$ and $\hat{S}_a$ based on Eq.~(\ref{eq:likelihood}), then apply bilinear upsampling $\text{Up}(\cdot)$ to match the anomaly mask $S \in \mathbb{R}^{H \times W}$:
\[
\scalebox{0.92}{$
L =
\text{Focal}\left( \text{Up}([\hat{S}_n, \hat{S}_a]), S \right)
+ \text{Dice}\left( \text{Up}([\hat{S}_n, \hat{S}_a]), S \right).
$}
\]

\textbf{Inference.} For each input image $x$, after computing visual outputs including local features $Z^M$ and global embeddings $\globalobject$ and $\globalspatial$ alongside detection and localization textual embeddings $\fixedfeatures$ and $\learnfeatures$ in forward pass, low-resolution anomaly map $\hat{S}_a$ is calculated, then bilinearly upsampled and smoothed with Gaussian filter to obtain final anomaly map $\hat{S}$:
\[
 \hat{S}_{a,(j,k)}= p_a(Z^M_{j,k}, \learnfeatures),\quad \hat{S}=Gaus(Up(\hat{S}_{a})).
\]
where $\hat{S}_{a,(j,k)}$ is anomaly score for each patch at location $(j, k)$. 

To calculate anomaly score \(\hat{y}\), we first compute the image-level score \(p_a(\globalspatial,\fixedfeatures)\) by comparing the global embedding \(\globalspatial\) with the abnormal/normal global prototypes \(g^a_f\) and \(g^n_f\) (collected in \(\fixedfeatures\)). We then inject strongest local evidence via the maximum pixel-level anomaly score \(\max(\hat{S}_a)\). The final score is
\[
\hat{y}=p_a(\globalspatial,\fixedfeatures)+\max(\hat{S}_a).
\]

%% file: sec/experiments.tex
\section{Experiments}
\label{sec:exp}
We evaluate on 14 diverse real-world anomaly detection datasets covering industrial defects and medical abnormalities, including MVTec-AD~\cite{bergmann2019mvtec}, VisA~\cite{zou2022spot-visa}, DTD-Synthetic~\linebreak\cite{aota2023zero-dtd}, SDD~\cite{tabernik2020segmentation-sdd}, BTAD~\cite{mishra2021vt-btad}, DAGM~\cite{wieler2007weakly-dagm}, and MPDD~\cite{jezek2021deep-mpdd} for industrial and ISIC~\cite{gutman2016skin}, CVC-ColonDB~\cite{tajbakhsh2015automated-colondb}, CVC-ClinicDB~\cite{bernal2015wm-clinicdb}, TN3K~\cite{gong2021multi-tn3k}, BrainMRI~\cite{BrainMRI}, HeadCT~\cite{HeadCT}, and BR35H~\cite{Br35h} for medical domain. Following prior works~\cite{zhou2023anomalyclip, cao2024adaclip}, we train on MVTec and test generalization on the others; for MVTec evaluation, we use VisA. In both cases, the categories are distinct and non-overlapping with those in the other datasets.
For metrics, we follow previous studies~\cite{chen2023april,zhou2023anomalyclip}, using AUROC, AP, and F1-max for image-level and AUROC, AUPRO, and F1-max for pixel-level, and report the mean over categories as the dataset-level score.

\input{tables/primary}

\input{tables/ablations}

\textbf{Implementation Details.}
We use the publicly available TIPS-L/14 HR~\cite{maninis2024tips} as our default backbone. It has 487.1M parameters, comparable to the CLIP-L/14@336 (428M) used in prior works. We use 8 learnable tokens for each of $T^{\mathrm{n}}$ and $T^{\mathrm{a}}$. For optimizer, we use Adam with a learning rate of 0.001 and \texttt{betas=(0.5, 0.999)}, and train for 2 epochs with a batch size of 8. The random seed for all algorithms is set to 111 to ensure reproducibility. Input images to the vision tower are resized to 518×518 and undergo the same normalization during both training and inference. Experiments are conducted using PyTorch 2.5 on an NVIDIA A6000 GPU.


\newcommand{\strong}[1]{#1\%}

\subsection{Quantitative Evaluations}
\textbf{Industrial Evaluation.}
We evaluate \methodname against three CLIP-based baselines~\cite{zhou2023anomalyclip,cao2024adaclip,chen2023april} on seven industrial datasets; results are summarized in Table~\ref{tab:main_results}. Averaged across datasets, \methodname improves image-level AUROC, AP, and F1-max by \strong{+2.3}, \strong{+3.9}, and \strong{+1.1} over prior work, attaining the top result on most datasets. Notably, these gains are achieved without complex and computationally costly adaptation tricks. At the pixel level, \methodname delivers even larger improvements in AUROC, AUPRO, and F1-max of \strong{+2.0}, \strong{+6.9}, and \strong{+1.5}, indicating stronger localization while maintaining a lightweight design.

\textbf{Medical Evaluation.}
To assess cross-domain generalization beyond industrial data, we evaluate \methodname on seven medical datasets; results are summarized in Table~\ref{tab:main_results}. At the pixel level, \methodname improves AUROC, AUPRO, and F1-max by a remarkable margin of \strong{+3.2}, \strong{+4.4}, and \strong{+5.3} on average while remaining competitive at image-level, indicating broader understanding and strong zero-shot generalization across challenging medical benchmarks.

\subsection{Qualitative Evaluations}
\begin{figure}[t]    
    \centering
    \includegraphics[width=1\columnwidth]{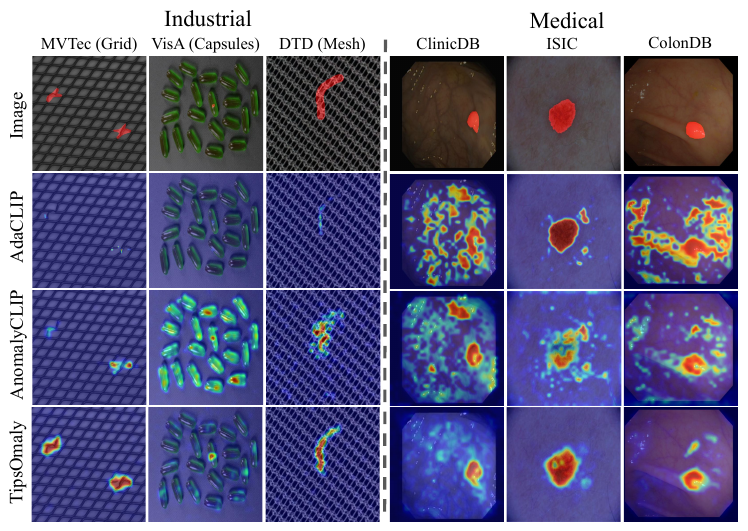} 
    \caption{\textbf{Qualitative results.} Anomalies colored in original images. \methodname yields cleaner localization with fewer false positives.}   
    \label{fig:qualitative}
\vspace{-1.2em}
\end{figure}

The qualitative comparison in Fig.~\ref{fig:qualitative} shows that our method produces cleaner anomaly maps with sharply reduced false positives and more complete coverage of true anomalous regions. In contrast, competing approaches either over-highlight normal areas or miss critical defect regions.

\vspace{-1em}
\subsection{Ablation Studies}
In following ablations we evaluate TIPS and highlight effect of each introduced strategies to address the TIPS challenges. 

\textbf{Decoupled Prompting.}
Table~\ref{table:decouple_prompt} compares fixed textual prompts, learnable prompts, and a decoupled design. Fixed prompts favor image-level classification but collapse on segmentation, whereas learnable prompts excel at pixel-level localization yet underperform in classification. \emph{Decoupling} resolves this inconsistency: we use fixed prompts for image-level scoring and learnable prompts for pixel-level maps, retaining strong performance at both levels.

\textbf{Learnable Prompts Objective.}
Table~\ref{table:learn_object} examines global vs.\ local losses for learning prompts. A \emph{local} loss alone yields the best pixel-level results. Adding a \emph{global} classification loss improves image-level performance slightly but reduces pixel-level accuracy, and still trails the image-level results obtained with fixed prompts denoting an existing \emph{distributional gap} between global and local embeddings. Hence, we rely on fixed prompts for image-level detection and optimize learnable prompts with the local loss for segmentation.

\textbf{Image-level Strategies.}
Table~\ref{table:imagelevel_strategy} assesses the discriminative ability of \(g^i_o\) and \(g^i_s\) for image-level anomaly detection, along with aggregation strategies (mean vs.\ max), with \(p_a(g^i_s,G_f)\) yielding the best discrimination. Finally, augmenting the global score with the strongest local evidence, \(+\max(\hat{S}_a)\), produces a net gain on the ten image-level datasets reported in Table~\ref{tab:main_results}, though this is not uniformly reflected by the two ablation sets. 

\section{Conclusion}
\vspace{-0.5em}
\label{sec:conclusion}
We investigated whether a spatially aware text–image backbone can deliver robust zero-shot anomaly detection without the intensive CLIP modifications common in prior work, demonstrating that TIPS can achieve this with a simple framework and fewer constraints. We showed a distributional gap between global and local visual embeddings in TIPS, which we bridged with decoupled prompting, resulting in the method \methodnamec which delivers state-of-the-art image- and pixel-level results across industrial and medical benchmarks, outperforming complex CLIP adaptations.
\vspace{-0.8em}

%% file: tables/primary.tex
\newcommand{\topone}[1]{\textbf{#1}}
\newcommand{\secondbest}[1]{\underline{#1}}

\definecolor{industrialColor}{RGB}{65, 105, 225}
\definecolor{medicalColor}{RGB}{0, 128, 128} 
\captionsetup[table]{skip=2pt}

\begin{table*}[t!]
\centering
\captionsetup{font=small}
\caption{\textbf{Comparison of ZSAD methods across industrial and medical domains.} 
The best performance is highlighted in \textbf{bold}, while the second-best is \underline{underlined}. Our method consistently outperforms baselines across different domains, showing strong zero-shot generalization.} 
\label{tab:main_results}

\begin{minipage}{0.48\linewidth}
\centering
\scriptsize 
\caption*{Image-level $\uparrow$ (AUROC, AP, F1-max)}
\resizebox{\linewidth}{!}{%
\begin{tabular}{@{}ccccc@{}}
\toprule[1.5pt]
Dataset & VAND~\cite{chen2023april} & AnomalyCLIP~\cite{zhou2023anomalyclip} & AdaCLIP~\cite{cao2024adaclip} & \textbf{Ours} \\
\midrule

\textcolor{industrialColor}{MVTec} &
    (86.1, 93.5, 88.9) & 
    (\secondbest{91.5}, \topone{96.2}, \secondbest{92.7}) &
    (89.2, 95.7, 90.6) &  
    (\topone{93.4}, \secondbest{96.1}, \topone{92.9}) \\
\textcolor{industrialColor}{VisA} &
    (78.0, 81.4, 80.7) &
    (82.1, \secondbest{85.4}, 80.4) &
    (\secondbest{85.8}, 79.0, \secondbest{83.1}) & 
    (\topone{87.7}, \topone{90.9}, \topone{84.8}) \\ 
\textcolor{industrialColor}{MPDD} & 
    (73.0, \secondbest{80.2}, 76.0) &
    (\topone{77.0}, \topone{82.0}, \secondbest{80.4}) &
    (\secondbest{76.0}, \secondbest{80.2}, \topone{82.5}) & 
    (74.7, 76.8, 79.7) \\
\textcolor{industrialColor}{BTAD} &
    (73.6, 68.6, 82.0) &
    (88.3, 87.3, 83.8) &
    (\secondbest{88.6}, \secondbest{93.8}, \secondbest{88.2}) & 
    (\topone{95.0}, \topone{95.5}, \topone{91.4}) \\
\textcolor{industrialColor}{KSDD} &
    (79.8, 71.4, 85.2) &
    (84.7, 80.0, 82.7) & 
    (\secondbest{97.1}, \secondbest{89.6}, \topone{90.7}) &  
    (\topone{97.8}, \topone{94.7}, \secondbest{90.1}) \\
\textcolor{industrialColor}{DAGM} & 
    (94.4, 83.8, 91.8) &
    (97.5, \secondbest{92.3}, 90.1) &
    (\secondbest{99.1}, 88.5, \topone{97.5}) & 
    (\topone{99.7}, \topone{98.1}, \secondbest{96.9})  \\ 
\textcolor{industrialColor}{DTD} &
    (94.6, 95.0, \secondbest{96.8}) &
    (93.5, 97.0, 93.6) &
    (\secondbest{95.5}, \secondbest{97.3}, 94.7) & 
    (\topone{99.4}, \topone{99.7}, \topone{99.1}) \\ \cmidrule{1-5}
\textcolor{industrialColor}{Average} & 
    (81.6, 82.0, 85.9) &
    (87.8, 88.6, 87.2) & 
    (\secondbest{90.2}, \secondbest{89.2}, \secondbest{89.6}) &
    (\topone{92.5}, \topone{93.1}, \topone{90.7})  \\

\cmidrule{1-5}
\textcolor{medicalColor}{HeadCT} & 
   (89.2, 89.5, 82.1) &
   (\topone{93.4}, \topone{91.6}, \topone{90.8}) &
   (91.8, \secondbest{90.6}, 84.1) & 
   (\secondbest{92.7}, 89.4, \secondbest{87.4}) \\
\textcolor{medicalColor}{BrainMRI} &  
   (89.6, 91.0, 88.5) &
   (90.3, 92.2, \topone{90.2}) &
   (\topone{93.5}, \topone{95.6}, \secondbest{89.7}) & 
   (\secondbest{92.4}, \secondbest{95.2}, 88.3) \\
\textcolor{medicalColor}{Br35H} & 
   (91.4, 91.9, 84.2) &
   (\topone{94.6}, \topone{94.7}, \topone{89.1}) &
   (92.3, 93.2, 85.3) & 
   (\secondbest{93.2}, \secondbest{93.6}, \secondbest{85.6}) 
   \\ \cmidrule{1-5}
\textcolor{medicalColor}{Average} &  
    (90.1, 90.8, 84.9) &
    (\topone{92.8}, \secondbest{92.8}, \topone{90.0}) &
    (\secondbest{92.5}, \topone{93.1}, 86.4) &
    (\topone{92.8}, 92.7, \secondbest{87.1}) \\

\bottomrule[1.5pt]
\end{tabular}}
\end{minipage}
\hfill
\begin{minipage}{0.48\linewidth}
\centering
\scriptsize
\caption*{Pixel-level $\uparrow$ (AUROC, AUPRO, F1-max)}
\resizebox{\linewidth}{!}{%
\begin{tabular}{@{}ccccc@{}}
\toprule[1.5pt]
Dataset & VAND~\cite{chen2023april} & AnomalyCLIP~\cite{zhou2023anomalyclip} & AdaCLIP~\cite{cao2024adaclip} & \textbf{Ours} \\
\midrule
\textcolor{industrialColor}{MVTec} & 
    (87.6, 44.0, 39.8) &
    (\topone{91.1}, \secondbest{81.4}, 39.1) &
    (88.7, 37.8, \secondbest{43.4}) & 
    (\secondbest{90.9}, \topone{84.0}, \topone{43.8}) \\
\textcolor{industrialColor}{VisA} &
    (94.2, 86.8, \secondbest{32.3}) &
    (\secondbest{95.5}, \secondbest{87.0}, 28.3) &
    (\secondbest{95.5}, 72.9, \topone{37.7}) &
    (\topone{95.9}, \topone{88.2}, 31.5) \\
\textcolor{industrialColor}{MPDD} &
    (94.1, 83.2, 30.6) &
    (\topone{96.5}, \topone{88.7}, 34.2) &
    (\secondbest{96.1}, 62.8, \secondbest{34.9}) &  
    (95.9, \secondbest{86.5}, \topone{36.1}) \\
\textcolor{industrialColor}{BTAD} &
    (60.8, 25.0, 38.4) &
    (\secondbest{94.2}, \secondbest{74.8}, 49.7) &
    (92.1, 20.3, \secondbest{51.7}) & 
    (\topone{96.7}, \topone{84.7}, \topone{56.2}) \\
\textcolor{industrialColor}{KSDD} &
    (79.8, 65.1, \secondbest{56.2}) &
    (90.6, \secondbest{67.8}, 51.3) &
    (\secondbest{97.7}, 33.8, 54.5) &
    (\topone{99.5}, \topone{97.9}, \topone{58.4}) \\
\textcolor{industrialColor}{DAGM} & 
    (82.4, 66.2, 57.9) &
    (\secondbest{95.6}, \secondbest{91.0}, \secondbest{58.9}) &
    (91.5, 50.6, 57.5) &  
    (\topone{97.2}, \topone{94.3}, \topone{65.3}) \\
\textcolor{industrialColor}{DTD} &
    (95.3, 86.9, \topone{72.7}) &
    (\secondbest{97.9}, \secondbest{92.3}, 62.2) &
    (\secondbest{97.9}, 72.9, \secondbest{71.6}) & 
    (\topone{99.3}, \topone{96.0}, 70.3) \\ \cmidrule{1-5}
\textcolor{industrialColor}{Average} &
    (84.9, 65.3, 46.8) &
    (\secondbest{94.5}, \secondbest{83.3}, 46.2) &
    (94.2, 50.1, \secondbest{50.2}) &
    (\topone{96.5}, \topone{90.2}, \topone{51.7}) \\ 

    \cmidrule{1-5}
\textcolor{medicalColor}{ISIC} &
       (89.5, 77.8, 71.5) & 
       (89.7, \secondbest{78.4}, 70.6) &
       (\secondbest{90.3}, 54.7, \secondbest{72.6}) & 
       (\topone{94.0}, \topone{87.0}, \topone{80.2}) \\
\textcolor{medicalColor}{ColonDB}  &
       (78.4, 64.6, 29.7) & 
       (81.9, \secondbest{71.3}, \secondbest{37.3}) &
       (\secondbest{82.6}, 66.0, 36.1) &
       (\topone{84.6}, \topone{75.0}, \topone{41.3}) \\
\textcolor{medicalColor}{ClinicDB} &  
       (80.5, 60.7, 38.7) & 
       (\secondbest{82.9}, \secondbest{67.8}, \secondbest{42.1}) &
       (82.8, 66.4, 40.9) & 
       (\topone{87.9}, \topone{77.4}, \topone{51.1}) \\
\textcolor{medicalColor}{TN3K} &
       (73.6, 37.8, 35.6) & 
       (\secondbest{81.5}, \topone{50.4}, \topone{47.9}) &
       (76.8, 34.0, 40.7) & 
       (\topone{82.4}, \secondbest{46.0}, \secondbest{46.5}) \\ \cmidrule{1-5}
\textcolor{medicalColor}{Average} & 
       (80.5, 60.2, 43.9) &
       (\secondbest{84.0}, \secondbest{67.0}, \secondbest{49.5}) &
       (83.1, 55.3, 47.6) &
       (\topone{87.2}, \topone{71.4}, \topone{54.8}) \\
\bottomrule[1.5pt]
\end{tabular}}
\end{minipage}
\end{table*}

%% file: tables/ablations.tex
\newcommand{\ablIndDataset}{MVTec}

\newcommand{\cmark}{\ding{51}} 
\newcommand{\xmark}{\ding{55}} 

\captionsetup{font=small}
\begin{table*}[htbp]
    \caption{\textbf{Ablation study of key components.} Performance is reported at image-level (AUROC, AP) and pixel-level (AUROC, AUPRO) on MVTec and VisA. Higher values indicate improved performance. \colorbox{blue!10}{Default configuration} is highlighted.}
    \centering
    \setlength{\tabcolsep}{3pt}

    {\scriptsize 
    \hspace*{-0.3cm}
    \begin{tabular}{ccc} 
        \begin{subtable}[t]{0.35\textwidth} 
            \centering
            \input{tables/abl-prompt}
        \end{subtable}
        &
        \begin{subtable}[t]{0.35\textwidth}
            \centering
            \input{tables/abl-seg-cls-both}
        \end{subtable}
        &
        \begin{subtable}[t]{0.26\textwidth}
            \centering
            \input{tables/abl-mean-max-obj-spat}
        \end{subtable}
    \end{tabular}


    \vspace{-1em} 
    } 

    \label{tab:ablation}
    \vspace{-1em}
\end{table*}

%% file: tables/abl-prompt.tex
\centering
\setlength{\tabcolsep}{2pt}
\caption{\textbf{Prompting approaches effect.}}

{\begin{tabular}{c cc cc}
    \toprule
      & \multicolumn{2}{c}{Image-level} &  \multicolumn{2}{c}{Pixel-level}  \\ 
     \cmidrule(r){2-3}
     \cmidrule(l){4-5}
    Prompts & \textbf{\ablIndDataset} & \textbf{VisA} & \textbf{\ablIndDataset} & \textbf{VisA} \\
    \midrule
    Fixed          & (\topone{93.4}, \topone{96.1}) & (\topone{87.7}, \topone{90.9}) & (55.4, 24.6) & (43.1, 11.5)\\
    Learned          & (84.4, 92.9) & (67.1, 72.9) & (\topone{90.9}, \topone{84.0}) & (\topone{95.9}, \topone{88.2})\\
    \rowcolor{blue!10}
    Decoupled         & (\topone{93.4}, \topone{96.1}) & (\topone{87.7}, \topone{90.9}) & (\topone{90.9}, \topone{84.0}) & (\topone{95.9}, \topone{88.2})\\
    \bottomrule
\end{tabular}
\label{table:decouple_prompt}}

%% file: tables/abl-seg-cls-both.tex
\centering
\setlength{\tabcolsep}{2pt}
\caption{\textbf{Loss types effect in learning prompts.}}

{\begin{tabular}{c cc cc}
    \toprule
      & \multicolumn{2}{c}{Image-level} &  \multicolumn{2}{c}{Pixel-level}  \\ 
     \cmidrule(r){2-3}
     \cmidrule(l){4-5}
    Loss & \textbf{\ablIndDataset} & \textbf{VisA} & \textbf{\ablIndDataset} & \textbf{VisA} \\
    \midrule
    global          & (\topone{94.2}, \topone{97.2}) & (\topone{81.4}, \topone{84.4}) & (73.9, 53.1) & (63.0, 36.7)\\
    local          & (84.4, 92.9) & (67.1, 72.9) & \cellcolor{blue!10}(\topone{90.9}, \topone{84.0}) & \cellcolor{blue!10}(\topone{95.9}, \topone{88.2})\\
    both         & (\secondbest{93.6}, \secondbest{97.0}) & (\secondbest{78.9}, \secondbest{82.6}) & (\secondbest{88.6}, \secondbest{77.4}) & (\secondbest{94.4}, \secondbest{84.9})\\
    \bottomrule
\end{tabular}
\label{table:learn_object}}

%% file: tables/abl-mean-max-obj-spat.tex
\centering
\setlength{\tabcolsep}{2pt}
\vspace{-2.5em}
\caption{\textbf{Image-level strategies effect.}}
{\begin{tabular}{c l cc cc}
    \toprule
      & & \multicolumn{2}{c}{Image-level}  \\ 
     \cmidrule(r){3-4}
    St. & Equation & \textbf{\ablIndDataset} & \textbf{VisA} \\
    \midrule
    S1 & $p_a(\globalobject,\fixedfeatures)$ & (93.5, 96.2) & (82.9, 85.5) \\
    S2 & $p_a(\globalspatial,\fixedfeatures)$ & (\topone{95.3}, \topone{97.6}) & (83.4, 85.8) \\
    S3 & $\max(S1, S2)$ & (\secondbest{95.0}, \secondbest{97.2}) & (\secondbest{83.7}, \secondbest{86.3}) \\
    S4 & $\text{mean}(S1, S2)$ & (94.4, 96.5) & (82.9, 85.8) \\
    \rowcolor{blue!10}S5 & $S2+\max(\hat{S}_a)$ & (93.4, 96.1) & \cellcolor{blue!10}(\topone{87.7}, \topone{90.9}) \\
    \bottomrule
\end{tabular}
\label{table:imagelevel_strategy}}

%% file: sec/appendix.tex
\newcommand{\tO}[1]{\topone{#1}}
\newcommand{\sB}[1]{\secondbest{#1}}

\subsection{Datasets}
We evaluate on 14 real-world anomaly detection datasets spanning industrial defects and medical abnormalities.
Industrial benchmarks include MVTec AD, VisA, DTD-Synthetic, SDD, BTAD, DAGM, and MPDD, which emphasize texture and structural defects in manufactured settings. Medical benchmarks comprise ISIC, CVC-ColonDB, CVC-ClinicDB, TN3K, BrainMRI, HeadCT, and BR35H, covering dermoscopic images, colonoscopy, thyroid ultrasound, and brain imaging (CT and MRI). Following prior work, we train on MVTec AD and assess generalization to the remaining datasets; to evaluate performance on MVTec itself, we train on VisA, which has non-overlapping categories. Table \ref{tab:datasets_details} reports per-dataset statistics (size, application, modality) and anomaly patterns. Note that some datasets are suitable only for localization or only for detection—for example, sets with only abnormal images and masks (such as ISIC) or sets lacking pixel-level annotations.

\input{tables/appendix_tables/datasets}

\subsection{Baseline Models}
We compare \methodname\ against three state-of-the-art ZSAD approaches on the 14 datasets. Reported performance numbers are taken from the original papers when available; otherwise, we reproduce results with the authors’ code—or, when that is unavailable, with widely used reference implementations—using default settings. Below we summarize each method and our reporting protocol.

\begin{itemize}
    \item \textbf{AdaCLIP}~\cite{cao2024adaclip}: Adapts CLIP via learnable deep tokens in both vision and text encoders, combining static (shared) and dynamic (image-specific) prompts. We use paper-reported values for all metrics except AUPRO and Image-AP, which are reproduced with the official code and default hyperparameters.

    \item \textbf{AnomalyCLIP}~\cite{zhou2023anomalyclip}: Employs object-agnostic learnable prompts (a generic \texttt{[object]} token) to highlight anomalous regions without category-specific tuning. All metrics are from the paper except image- and pixel-level F1-max, which we reproduce with the official 
    codebase and defaults.

    \item \textbf{VAND}~\cite{chen2023april}: Fine-tunes the vision encoder with linear projections using an auxiliary training set; prompting follows WinCLIP/AnoVL. MVTec-AD and VisA metrics are taken from the paper; for datasets lacking official numbers, we reproduce results with the authors’ default settings and official repository.
\end{itemize}

\subsection{Comparison of CLIP and TIPS variants}
Table~\ref{tab:variants_tips_clip} compares model sizes across TIPS and CLIP families (embedding dimension and parameter count). Most prior ZSAD works evaluate with OpenCLIP ViT-L/14@336px, so we capacity-match our backbone by using TIPS-L/14 HR which lies in the same size tier and enables a fair comparison.

\input{tables/appendix_tables/abl-tips-clips}

\subsection{More ablations}

\subsubsection{Effect of TIPS's variants as backbone}
Table~\ref{table:abl-model-size} evaluates TIPS backbones of different sizes. We observe consistent gains as the backbone scales from S to L. Overall, L/14 HR—which is comparable in size to the default backbone in prior works, CLIP ViT-L/14 (Table~\ref{tab:variants_tips_clip})—performs better across datasets and evaluation levels. The g family is generally the next best, though the low-resolution training setup tends to hurt localization. When compute is constrained, B/14 HR offers a promising accuracy–efficiency trade-off, while S/14 HR remains feasible for strictly resource-limited settings.

\subsubsection{Effect of domain-aware medical prompts}
We performed an ablation study to examine how fixed prompt design influences performance in more specialized domains. Generic states (Sec. 2.3) such as “damaged” or “with defect” can represent common anomaly patterns in natural or industrial images such as broken, cracked, or deformed objects. We evaluate whether such general descriptions can also transfer effectively to medical imaging, where anomalies follow domain-specific morphological and pathological patterns.

To investigate this, we designed a set of medically aware prompts and compared their anomaly-detection performance with that of generic prompts. The normal set provides low-level descriptors of healthy tissue conditions and the abnormal set incorporates terms such as diseased and phrases denoting localized structural changes, including irregular area, and uneven texture, as listed in Table \ref{tab:medical_prompts}. Each prompt begins with a contextual medical template (e.g., “a medical image of a { }”, “a diagnostic scan of a { }”) to explicitly anchor the textual representation to the medical imaging domain.

\begin{table}[h]
\centering
\caption{\textbf{Medical normal and abnormal states templates}}
\label{tab:medical_prompts}
\renewcommand{\arraystretch}{1.2}
\begin{tabular}{p{4cm} p{3.5cm}}
\toprule
\textbf{Normal} & \textbf{Abnormal} \\
\midrule

\begin{tabular}[t]{l}
normal \{\} \\
intact \{\} \\
\{\} with uniform structure \\
\{\} showing clear tissue \\
\{\} with normal anatomy \\
\{\} showing no distortion \\
\{\} with symmetric appearance \\
\{\} looking normal \\
\{\} with even texture \\
\{\} with regular shape
\end{tabular}
&
\begin{tabular}[t]{l}
abnormal \{\} \\
\{\} with spot \\
\{\} with abnormality \\
diseased \{\} \\
\{\} showing distortion \\
\{\} with irregular area \\
\{\} with irregular shape \\
\{\} with uneven texture
\end{tabular}
\\

\bottomrule
\end{tabular}
\end{table}

Table \ref{table:abl-medical-prompts} summarizes the results of evaluating generic prompts versus medically aware prompts on the HeadCT and BrainMRI datasets. Medically relevant descriptions consistently yield higher performance across all datasets and metrics, indicating that fixed prompts benefit from domain-aware formulations, which steer the textual representation toward better alignment with the corresponding visual patterns. We adopt the medically aware prompts as the default for all medical classification datasets.

\begin{table}[htbp]
\centering
\setlength{\tabcolsep}{6pt}
\caption{\textbf{Effect of domain-aware prompts for medical datasets.} Image-level results are reported on HeadCT and BrainMRI. Default row is highlighted.}
\caption*{Image-level $\uparrow$ (AUROC, AP, F1-max)}
\begin{tabular}{c c c}
\toprule
Prompt & \textbf{HeadCT} & \textbf{BrainMRI}\\
\midrule
Generic       & (82.9, 77.6, 78.6) &  (87.7, 90.1, 85.1) \\
\rowcolor{blue!10}
Medical  & (\topone{92.7}, \topone{89.4}, \topone{87.4}) & (\topone{92.4}, \topone{95.2}, \topone{88.3}) \\

\bottomrule
\end{tabular}
\label{table:abl-medical-prompts}
\end{table}

\vspace{-0.5em}
\subsubsection{Effect of utilizing Intermediate features}
We evaluated the effect of aggregating dense output features from the intermediate layers of TIPS-L/14-HR in Table~\ref{table:abl-intermediate-layers}. Unlike CLIP, which benefits from intermediate-layer aggregation in dense tasks by recovering spatial detail that is attenuated in deeper blocks, TIPS shows no meaningful gain; in fact, incorporating shallower layers tends to degrade performance. Notably, in the second row, combining layer 18 features produces minor gains on a few metrics, yet the overall performance still declines.
\input{tables/appendix_tables/abl-inter-layers}

\vspace{-0.5em}
\subsubsection{Effect of number of learnable tokens}
\input{tables/appendix_tables/abl-model-size}
\input{tables/appendix_tables/abl-siglip2-performance}

\begin{figure*}[tbhp]
    \centering
    \includegraphics[width=\linewidth]{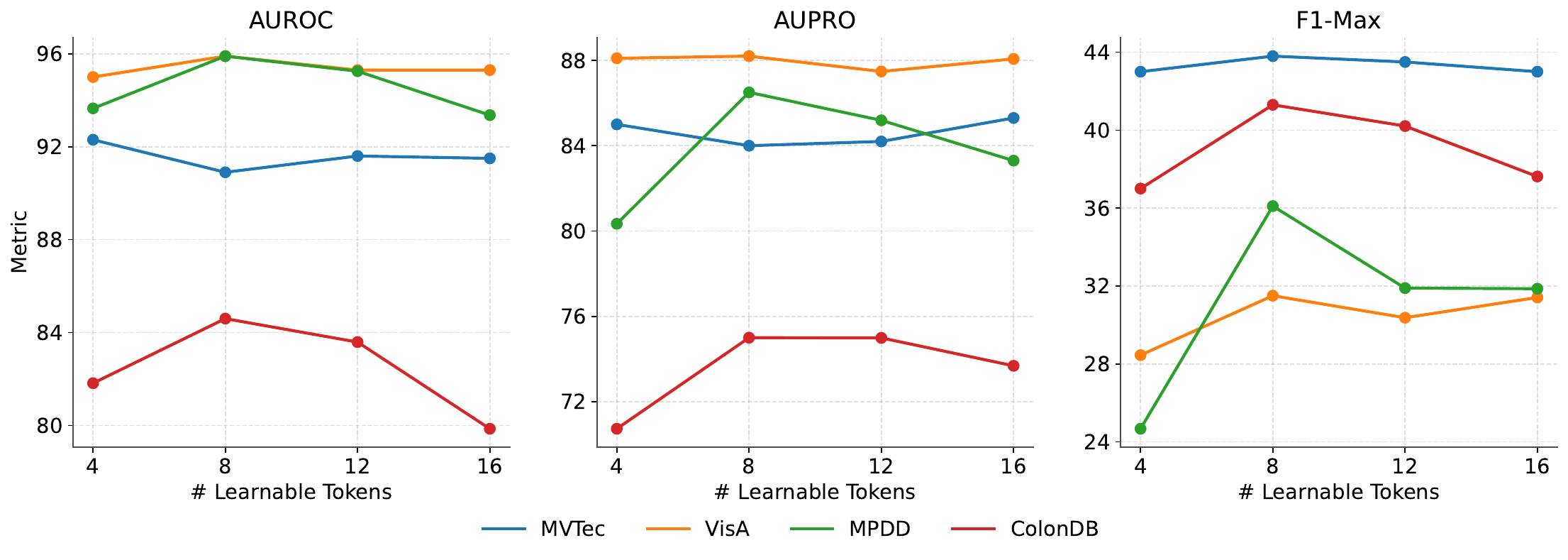}
    \vspace{-2.2em}
    \caption{\textbf{Ablation over number of learnable tokens.} The performance over various metrics and datasets suggests longer learnable prompts may overfit to the source domain and lose their ability to generalize effectively. Using 8 learnable tokens yields the best overall results.}
    \label{fig:nprompts_ablation}
\end{figure*}
The effect of the learnable-prompt length is shown in Figure~\ref{fig:nprompts_ablation}. As the number of parameters increases, the overall performance shows little to no improvement and even declines, particularly on domains farther from the training data such as MPDD and CVC-ColonDB, indicating reduced generalization with longer learnable prompts. Consequently, we select a length of 8 for learnable prompts, as it yields the most balanced performance across the evaluated settings.

\subsubsection{Effect of SigLIP2 as backbone}
SigLIP2~\cite{tschannen2025siglip} is another state-of-the-art vision--language encoder proposed to improve spatial awareness. Unlike CLIP and TIPS, which use contrastive loss, SigLIP2 utilizes a sigmoid loss to learn a shared image--text representation space. We replace TIPS with SigLIP2 in this method and report its anomaly detection and localization capabilities in Table~\ref{table:abl-siglip2}. As shown, a relatively similar trend to TIPS can be observed: the best localization performance is achieved by training only with the local loss, while the best overall classification performance is obtained with fixed prompts (No Learning). However, under this framework, SigLIP2 yields poorer results than TIPS, especially in localization, making it a less favorable backbone than TIPS.

\subsubsection{Category-level qualitative results}
To analyze the category-level behavior of \methodname and provide a more intuitive understanding of its pixel-level output, we visualize localization results for 10 anomalous images from each of the Capsule, Hazelnut, Metal Nut, Screw, and Wood categories in MVTec-AD (Figures~4--8), highlighting its ability to precisely localize diverse defect patterns.

\newlength{\captpfigspace}
\setlength{\captpfigspace}{-1.7em}

\newlength{\figfigspace}
\setlength{\figfigspace}{-1em}

\begin{figure*}[hbtp]
    \centering
    \includegraphics[width=\linewidth]{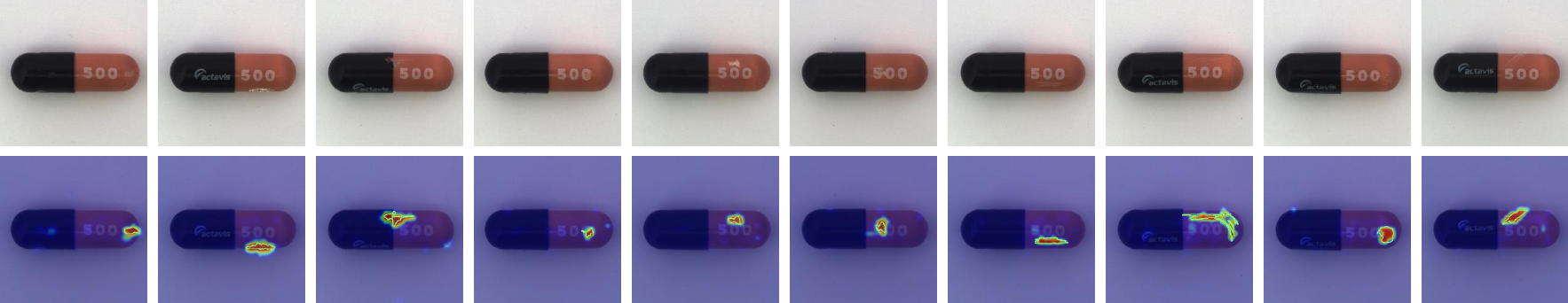}
    \vspace{\captpfigspace}
    \caption{\textbf{Category-level localizations for the object capsule in the MVTec-AD.} Ground-truth anomalous regions are outlined in green.}
    \label{fig:capsule}
\vspace{\figfigspace}
\end{figure*}

\begin{figure*}[hbtp]
    \includegraphics[width=\linewidth]{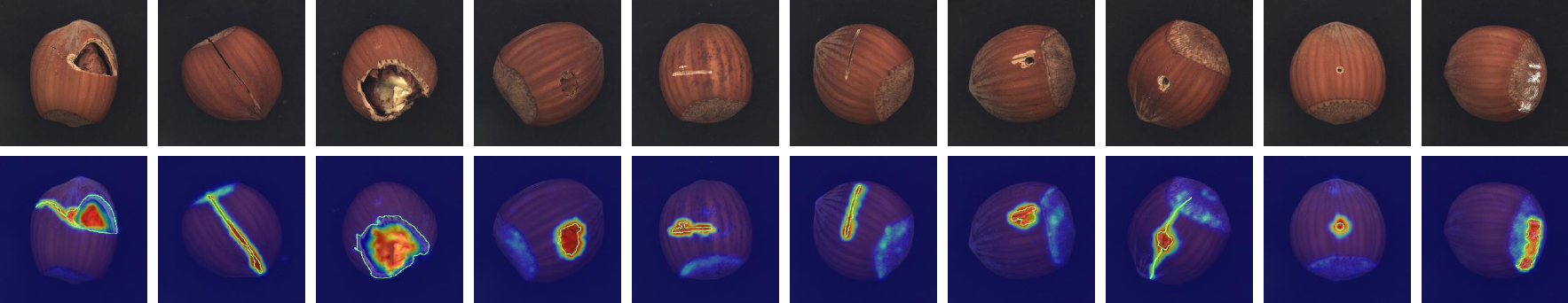}
    \vspace{\captpfigspace}
    \caption{\textbf{Category-level localizations for the object hazelnut in the MVTec-AD.} Ground-truth anomalous regions are outlined in green.}
    \label{fig:hazelnut}
\vspace{\figfigspace}
\end{figure*}

\begin{figure*}[hbtp]
    \includegraphics[width=\linewidth]{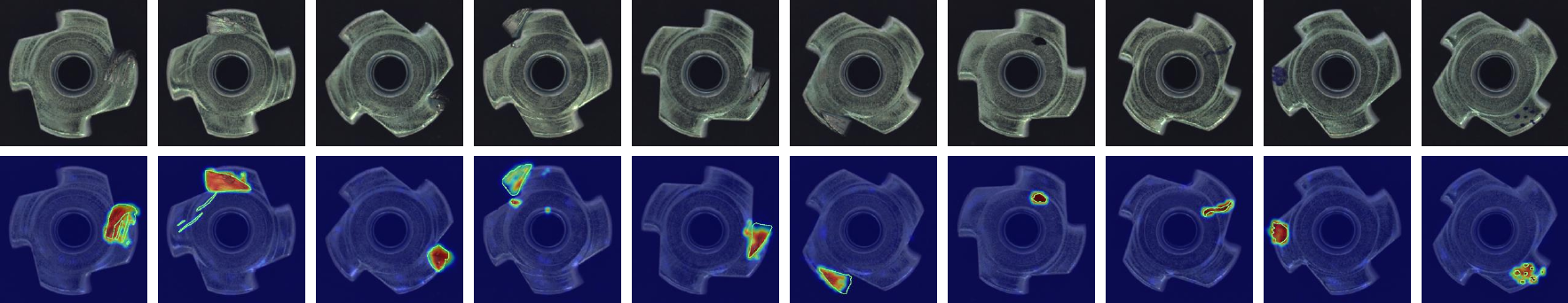}
    \vspace{\captpfigspace}
    \caption{\textbf{Category-level localizations for the object metal\_nut in the MVTec-AD.} Ground-truth anomalous regions are outlined in green.}
    \label{fig:metal_nut}
\vspace{\figfigspace}
\end{figure*}

\begin{figure*}[hbtp]
    \includegraphics[width=\linewidth]{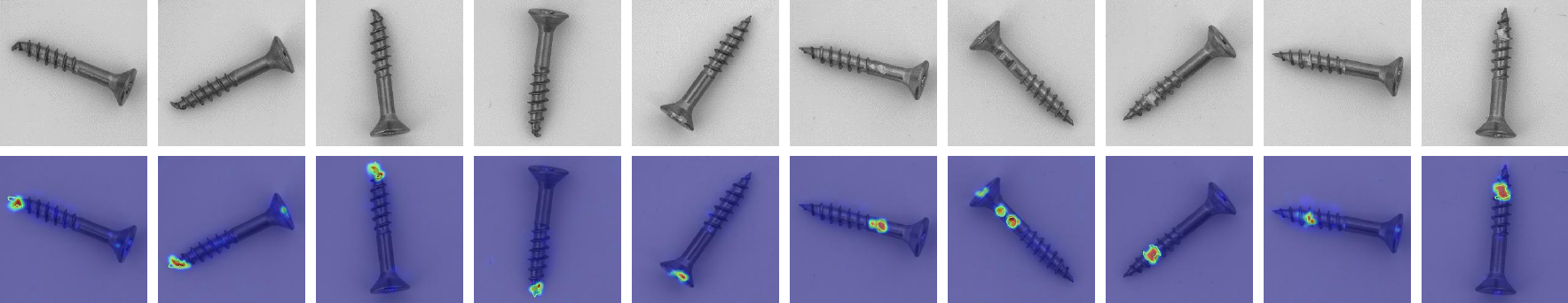}
    \vspace{\captpfigspace}
    \caption{\textbf{Category-level localizations for the object screw in the MVTec-AD.} Ground-truth anomalous regions are outlined in green.}
    \label{fig:screw}
\vspace{\figfigspace}
\end{figure*}

\begin{figure*}[hbtp]
    \includegraphics[width=\linewidth]{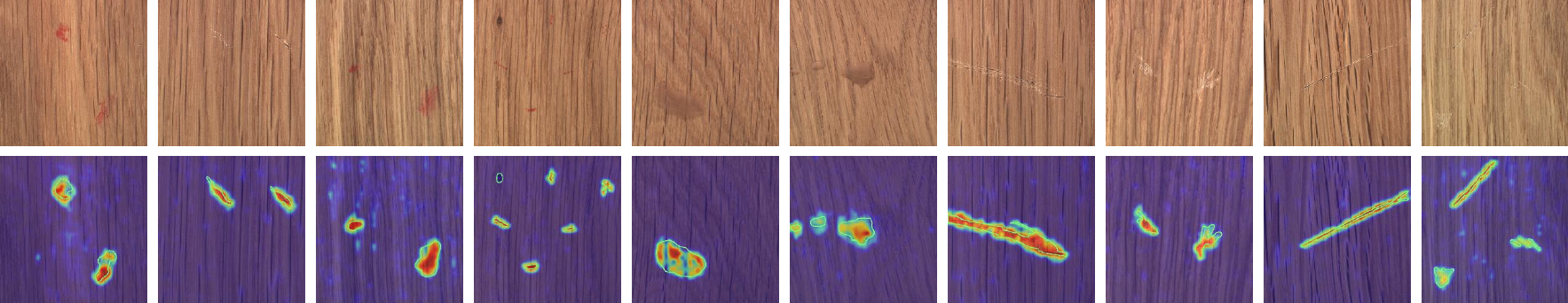}
    \vspace{\captpfigspace}
    \caption{\textbf{Category-level localizations for the object wood in the MVTec-AD.} Ground-truth anomalous regions are outlined in green.}
    \label{fig:wood}
\vspace{\figfigspace}
\end{figure*}

%% file: tables/appendix_tables/datasets.tex
\definecolor{industrialColor}{RGB}{65, 105, 225}
\definecolor{medicalColor}{RGB}{0, 128, 128} 
\newlength{\colwidth}
\settowidth{\colwidth}{normal \& anomalous}

\begin{table}[htbp]
    \centering
    \caption{\textbf{Numerical details of the utilized datasets.} $|C|$ denotes the number of object categories in each dataset.}
    \label{tab:datasets_details}
    \resizebox{\linewidth}{!}{
    \begin{tabular}{ccccc}
    \toprule 
    Dataset & Modalities & $|C|$ & \# Normal \& Anomalous & Detection\\  
    \midrule
    \textcolor{industrialColor}{MVTec-AD}\textsuperscript{3} & Photo & 15 & (467, 1258) & Industrial defect \\ 
    \cmidrule(lr){1-5}
    \textcolor{industrialColor}{VisA}\textsuperscript{3} & Photo & 12 & (962, 1200) & Industrial defect \\  
    \textcolor{industrialColor}{MPDD}\textsuperscript{3} & Photo & 6 & (176, 282) & Industrial defect \\  
    \textcolor{industrialColor}{BTAD}\textsuperscript{3} & Photo & 3 & (451, 290) & Industrial defect \\  
    \textcolor{industrialColor}{SDD}\textsuperscript{3} & Photo & 1 & (181, 74) & Industrial defect \\  
    \cmidrule(lr){1-5}
    \textcolor{industrialColor}{DAGM}\textsuperscript{3} & Photo & 10 & (6996, 1054) & Industrial defect \\  
    \textcolor{industrialColor}{DTD-Synthetic}\textsuperscript{3} & Photo & 12 & (357, 947) & Industrial defect \\  
    \midrule
    \textcolor{medicalColor}{ISIC}\textsuperscript{2} & Photo & 1 & (0, 379) & Skin cancer \\  
    \cmidrule(lr){1-5}
    \textcolor{medicalColor}{CVC-ClinicDB}\textsuperscript{2} & Endo. & 1 & (0, 612) & Colon polyp \\  
    \textcolor{medicalColor}{CVC-ColonDB}\textsuperscript{2} & Endo. & 1 & (0, 380) & Colon polyp \\   
    \cmidrule(lr){1-5}
    \textcolor{medicalColor}{TN3K}\textsuperscript{2} & US & 1 & (0, 614) & Thyroid nodule \\  
    \cmidrule(lr){1-5}
    \textcolor{medicalColor}{HeadCT}\textsuperscript{1} & CT & 1 & (100, 100) & Brain tumor \\  
    \textcolor{medicalColor}{BrainMRI}\textsuperscript{1} & MRI & 1 & (98, 155) & Brain tumor \\  
    \textcolor{medicalColor}{Br35H}\textsuperscript{1} & MRI & 1 & (1500, 1500) & Brain tumor \\  
    \bottomrule
    \multicolumn{5}{l}{\footnotesize%
    Annotations: 
\textsuperscript{1} image-level only, 
\textsuperscript{2} pixel-level only, 
\textsuperscript{3} both levels.}
    \end{tabular}%
}
\end{table}

%% file: tables/appendix_tables/abl-tips-clips.tex
\begin{table}[htbp]
\centering
\setlength{\tabcolsep}{15pt} 
\caption{\textbf{Variants of TIPS vs. CLIP}}

\resizebox{\columnwidth}{!}{%
\begin{tabular}{ccc}
    \toprule
    Model Size & Embed. Dim. & \# Parameters \\
    \midrule
    TIPS-S/14 HR & 384 & 55.2M \\
    \midrule
    TIPS-B/14 HR & 768 & 195.3M \\
    CLIP ViT-B/16 & 512 & 150M \\
    CLIP ViT-B/32 & 512 & 151M \\
    \midrule
    TIPS-L/14 HR & 1024 & 487.1M \\
    CLIP ViT-L/14 & 768 & 428M \\
    \midrule
    TIPS-SO/14 HR & 1152 & 860.8M \\
    CLIP ViT-H/14 & 1024 & 986M \\
    \midrule
    TIPS-g/14 LR & 1536 & 1.5B \\
    TIPS-g/14 HR & 1536 & 1.5B \\
    CLIP ViT-g/14 & 1024 & 1.37B \\
    CLIP ViT-G/14 & 1280 & 2.54B \\
    \bottomrule
\end{tabular}
}
\label{tab:variants_tips_clip}
\end{table}

%% file: tables/appendix_tables/abl-inter-layers.tex
\begin{table}[hptb]
\centering
\setlength{\tabcolsep}{6pt}
\caption{\textbf{Ablation over intermediate features.} Image-/pixel-level results are reported on MVTec and VisA. Default row is highlighted.}
\caption*{Pixel-level $\uparrow$ (AUROC, AUPRO, F1-max)}
\begin{tabular}{c c c}
\toprule
Layer Combination & \textbf{MVTec} & \textbf{VisA} \\
\midrule
\rowcolor{blue!10}
\{24\}            & (\secondbest{90.9}, \secondbest{84.0}, \topone{43.8}) & (\topone{95.9}, \topone{88.2}, \topone{31.5}) \\
\{18, 24\}    & (\topone{91.1}, \topone{84.7}, \secondbest{43.0}) & (\secondbest{95.4}, \secondbest{88.0}, \topone{31.5}) \\
\{12, 18, 24\}   & (90.2, 84.2, 41.5) & (95.2, 87.2, 30.1) \\
\{6, 12, 18, 24\} & (87.6, 81.9, 40.6) & (95.0, 87.7, 29.5) \\
\bottomrule
\end{tabular}
\label{table:abl-intermediate-layers}
\end{table}

%% file: tables/appendix_tables/abl-model-size.tex
\begin{table*}[tbhp]
\setlength{\tabcolsep}{8pt}
\centering
\caption{\textbf{Ablation over TIPS variants.} Image- and pixel-level Performance are reported on MVTec and VisA.  Default row is highlighted.}
\begin{tabular}{c cc cc}
    \toprule
      & \multicolumn{2}{c}{\makecell{Image-level $\uparrow$ (AUROC, AP, F1-max)}} &  \multicolumn{2}{c}{\makecell{Pixel-level $\uparrow$ (AUROC, AUPRO, F1-max)}}  \\ 
     \cmidrule(r){2-3}
     \cmidrule(l){4-5}
    Model Size & \textbf{\ablIndDataset} & \textbf{VisA} & \textbf{\ablIndDataset} & \textbf{VisA} \\
    \midrule
    TIPS-S/14 HR & (87.3, 93.5, 90.1)                & (83.6, 86.9, 81.3)                & (89.6, 80.7, 38.6)                & (93.3, 82.8, 24.4)\\
    TIPS-B/14 HR & (91.8, \tO{96.3}, \sB{92.4})      & (\sB{84.2}, \sB{88.1}, \sB{82.0}) & (89.0, \tO{84.8}, 39.7)           & (93.1, 78.5, 25.6)\\
    \rowcolor{blue!10}
    TIPS-L/14 HR & (\tO{93.4}, \sB{96.1}, \tO{92.9}) & (\tO{87.7}, \tO{90.9}, \tO{84.8}) & (\tO{90.9}, \sB{84.0}, \tO{43.8}) & (\tO{95.9}, \sB{88.2}, \tO{31.5})\\
    TIPS-g/14 LR & (89.3, \sB{96.1}, 91.8)           & (83.9, 86.8, 81.4)                & (87.8, 79.1, 36.4)                & (95.1, \tO{88.4}, 29.0)\\
    TIPS-g/14 HR & (\sB{92.9}, 96.0, \tO{92.9})      & (82.0, 86.7, 80.4)                & (\sB{90.6}, 77.1, \sB{43.3})      & (\sB{95.3}, 82.0, \sB{29.8})\\
    \bottomrule
\end{tabular}
\label{table:abl-model-size}
\end{table*}

%% file: tables/appendix_tables/abl-siglip2-performance.tex
\begin{table*}[tbhp]
\setlength{\tabcolsep}{8pt}
\centering
\caption{\textbf{Effect of SigLIP2 as backbone.} Image- and pixel-level Performance are reported on MVTec and VisA.}
\begin{tabular}{c cc cc}
    \toprule
      & \multicolumn{2}{c}{\makecell{Image-level $\uparrow$ (AUROC, AP, F1-max)}} &  \multicolumn{2}{c}{\makecell{Pixel-level $\uparrow$ (AUROC, AUPRO, F1-max)}}  \\ 
     \cmidrule(r){2-3}
     \cmidrule(l){4-5}
    Loss & \textbf{\ablIndDataset} & \textbf{VisA} & \textbf{\ablIndDataset} & \textbf{VisA} \\
    \midrule
    No Learning & (\secondbest{88.7}, \secondbest{94.4}, \secondbest{90.1})                & (\tO{74.9}, \tO{80.5}, \tO{78.1})                & (47.3, 00.1, 06.6)                & (47.0, 00.0, 01.2)\\
    Local Loss & (60.4, 82.1, 83.8)      & (43.2, 54.5, 72.7) & (\tO{61.3}, \tO{31.0}, \tO{10.3})           & (\tO{56.9}, \tO{27.9}, \tO{02.2})\\
    Global Loss & (\tO{91.6}, \tO{96.6}, \tO{92.1})      & (71.4, 76.2, 76.1) & (48.9, 00.2, 06.8)           & (48.4, 00.0, 01.3)\\
    Both Losses & (67.1, 85.4, 84.3)      & (\secondbest{74.4}, \secondbest{79.0}, \secondbest{77.9}) & (\secondbest{57.5}, \secondbest{25.8}, \secondbest{09.6})           & (\secondbest{57.2}, \secondbest{26.3}, \secondbest{02.0})\\
    \bottomrule
\end{tabular}
\label{table:abl-siglip2}
\end{table*}